\documentclass[letterpaper]{article}
\pdfoutput=1
\usepackage{calc,amsmath,amssymb,amsfonts}
\usepackage[T1]{fontenc}
\usepackage{ae}
\usepackage{aecompl}
\usepackage[greek,english]{babel}
\usepackage{xcolor,fancyhdr}
\usepackage[top=0.5in, bottom=0.5in, hmargin=1in, includeheadfoot, head=0.5in, headsep=0.4in]{geometry}
\usepackage{array,supertabular,hhline,enumitem,hyperref}
\hypersetup{colorlinks=true,allcolors=blue,pdfauthor=Dasgupta; Sudip}
\usepackage[pdftex]{graphicx}
\usepackage{tikz}
\usepackage{ragged2e}
\definecolor{colori}{HTML}{A0A0A0}
\makeatletter
\newcommand\arraybslash{\let\\\@arraycr}
\makeatother
\usepackage{fancyhdr}
  \renewcommand\thepage{\arabic{page}}
\author{Dasgupta, Sudip}
\date{2025-06-13}
\begin{document}
\setlength{\parskip}{1em}
\setlength{\parindent}{0pt}

\clearpage
\begin{center}
    \hrule\vspace{0.5em}
\section{AI Agents-as-Judge: Automated Assessment of Accuracy, Consistency, Completeness and Clarity for Enterprise Documents}
    \vspace{0.5em}\hrule
\end{center}

\bigskip
\begin{center}
\textbf{Sudip Dasgupta} \hspace{2.7 cm}\textbf{Himanshu Shankar}\\
\hspace{0.0000000001 cm}Senior Manager  \hspace{ 2.7cm}Associate Director\\
\hspace{-0.3cm}Accenture\hspace{3.7cm}Accenture\\
\hspace{0.5cm}sudip.dasgupta77@gmail.com \hspace{0.5cm}shankarhimanshu@gmail.com\\
\end{center}
\bigskip

\subsection{\textbf{The Problem}}

This study is about finding out how good and reliable AI Agents are at automatically checking all kinds of business documents created by companies. Instead of only looking at one type of document or one area, the study looks at many types—like documents needed for government rules, company procedures, or any other business requirement.

The main question is: Can we trust these AI Agents to review business documents on their own, when the documents need to follow strict formats, use the correct terms and be factually correct?

\subsection{\textbf{The Objectives}}

\subsubsection{1. Can AI Do the Job Well?}

The study checks if using AI Agents really works for reviewing business documents. This includes seeing if the AI Agents can make sure documents match the right template (for example, as needed for government rules or business needs), contain accurate information, and use the right words for the industry.

\noindent\textit{Example:} Can the AI Agents make sure a company’s report follows a specific layout and uses correct technical words?

\subsubsection{2. Build a Flexible Review System}

The goal is to create a system that uses modern tools (like LangChain, CrewAI, TruLens, and Guidance) so that it can easily adjust to different types of business documents and changing quality needs.

\subsubsection{3. Compare AI and Human Reviews}

The study measures how well and how quickly the AI Agents review documents and compares these results to what human reviewers do. This helps find out where AI Agents do better and where humans are still needed.

\noindent\textit{Example:} Check if the AI Agents find mistakes in business documents faster or more accurately than people.

\subsubsection{4. Easy-to-Follow Instructions}

The study will create simple and practical steps for using AI Agents in document reviews. This includes advice on how to ask the AI the right questions, how to get organized answers, and how to keep improving the process by learning from real-life results.

\noindent\textit{Example:} Give clear templates for asking AI Agents to find important information in business documents, and advice on changing these questions as rules change. 

\section{\textbf{The Significance}}

\textbf{Saves Time and Work:}

Using AI Agents make the review of business documents much faster and easier, so people don’t have to spend so much time on it. It also helps reduce mistakes.

\noindent \textbf{Keeps Things Fair and Consistent:}

AI Agents review documents the same way every time, making the results fair and steady, unlike humans, who might do things differently each time.

\noindent \textbf{No Personal Bias:}

Because AI Agents use fixed rules, it doesn’t bring personal opinions into the review. This is very important in situations where fairness and accuracy really matter, like in healthcare or official paperwork.

\noindent \textbf{Works More Efficiently:}

Because AI Agents use fixed rules, it doesn’t bring personal opinions into the review. This is very important in situations where fairness and accuracy really matter, like in healthcare or official paperwork.

\noindent\textbf{Reusable AI System:}

The study will provide a step-by-step system that anyone—businesses or researchers—can use to bring AI into the review process for different types of documents, from healthcare forms to financial reports.

\noindent\textbf{Shows What AI Can't Do (Yet):}

The study will also point out where current AI tools still need human help, especially for very complex or sensitive documents.

\noindent \textit{Example: Some medical reports may need a doctor to review them, even if the AI Agents can check that the report follows the right format. }

\section{\textbf{Introduction}}

The growing use of powerful AI systems using AI Agents by calling Large Language Models (LLMs), like GPT-4 and Llama 2 and other tools, has changed the way companies review business documents. These AI Agents can be used for many jobs—like checking if a document meets rules, grading documents, or making sure technical details are correct. However, most research and real-world use focuses on documents that don’t have a fixed structure. There is still a big gap when it comes to reviewing highly structured, template-based business documents, which need each section to follow strict rules and business logic.

These kinds of business documents—such as those used for regulatory filings, internal process documentation, or other business requirements—need to be checked not just for accuracy and completeness, but also for correct layout, specific technical words, and how well they match related documents. Reviewing these by hand takes a lot of time and effort, can lead to mistakes, and is not always consistent. This is why companies need faster, automatic, and more reliable ways to check such documents.

Some open-source AI tools, like DeepEval and Evidently AI, help review text made by LLMs. They let you try different questions, track results, and look for mistakes. But these tools mostly work well for simple text or chatbot conversations. They are not designed for complex business documents that follow a strict format and need to meet specific business needs.

This study aims to solve this problem. It looks at how to build a flexible and detailed system, using AI Agents and tools like LangChain, CrewAI, TruLens, Guidance and others, to automatically and accurately review structured business documents created for enterprises. The goal is to create a process that is clear, section-by-section, and easy to check.

\section{\textbf{Novelty and Advantages}}

The invention described herein introduces a modular, multi-agent pipeline for the automated evaluation of enterprise business documents using AI Agents and other tools. Unlike existing document automation solutions, which typically focus on generic or unstructured texts, this system is specifically designed for highly structured, template-based enterprise documents that require section-by-section review.

 Key novel features include:
 
\begin{itemize}
    \item A multi-agent architecture that assigns specialized review tasks—such as template compliance, factual accuracy, and completeness—to dedicated agents, enabling parallel and expert-level evaluation of each document section.

    \item The use of orchestration frameworks that allow for flexible adaptation to any document structure or business template, supporting rapid integration with evolving enterprise requirements.

    \item An enforced, machine-readable output schema that ensures all evaluation results are standardized, auditable and easily integrated into downstream analytics or compliance workflows.

    \item A continuous monitoring and feedback loop, allowing for iterative improvement and bias mitigation based on real-world reviewer input.

    \item Scalability to process large volumes of documents across various formats and industries, far surpassing the manual review process in speed, accuracy, and consistency.
\end{itemize}

\section{\textbf{Comparison with Prior Art}}

Most existing enterprise document review systems utilize rule-based engines, template checkers, or simple text analytics tools and are primarily limited to static checks or unstructured text. Some recent solutions leverage AI for text classification or extraction but do not support dynamic, agent-based evaluation or adaptable orchestration across varied document templates.

Unlike these prior systems, the present invention:

\begin{itemize}
    \item Provides a modular, agent-driven pipeline where different evaluation criteria are handled by specialized agents, not by a monolithic engine.

    \item Supports real-time, section-by-section review and feedback, rather than batch or end-of-document analysis.

    \item Enforces structured outputs at every stage, enabling seamless auditability and integration with business intelligence tools.

    \item Integrates a continuous feedback mechanism, allowing both automated refinement and human-in-the-loop oversight, which are absent in most legacy tools.

    \item Is designed to be compatible with any AI model or agent framework, allowing rapid adoption of new AI technologies.
\end{itemize}

\section{Literature Review and Choosing Tools}

Recent studies indicate a growing trend in the use of AI agents for reviewing business documents, particularly in assessing accuracy, compliance, and technical soundness. Research demonstrates that, when provided with well-structured instructions, AI agents can sometimes match or even surpass human performance in identifying errors and ensuring adherence to established guidelines.

However, several challenges remain. AI agents can still produce inconsistent results, generate fabricated information (hallucinations), or exhibit bias. To address these issues, experts recommend breaking down evaluation instructions into clear, manageable steps, developing detailed rubrics, and maintaining continuous oversight of AI outputs. Open-source tools such as OpenAI Evals, DeepEval, and Evidently AI have emerged as valuable resources, facilitating the setup of rigorous evaluation processes and providing robust monitoring capabilities for AI-based assessments. 

\section{\textbf{Using LangChain, CrewAI, TruLens, and Guidance}}

Different tools help put these ideas into practice for reviewing structured business documents:

\textbf{LangChain: }Helps build systems that organize the review process step-by-step. For example, it can break a document into sections, check each part and combine the results.

\textbf{CrewAI:} Allows the work to be split among several “agents,” just like people in a review team. One agent might check if the document uses the right template, another checks for accuracy, and another makes sure the right words are used.

\textbf{TruLens:} Provides dashboards to monitor the AI Agent answers, watch for bias, and keep track of quality over time. This is important for documents where accuracy and fairness are critical.

\textbf{Guidance:} Makes sure the AI Agent answers are always given in a clear, set format (like tables or lists). This makes it easy for computers to read the results and for people to audit them later.

\section{\textbf{How Research and Tools Work Together}}

Here is how research advice matches up with real-world tools:
\begin{flushleft}
\tablefirsthead{\hline
\centering \textbf{Research Insight} &
\centering \textbf{Industry Tool Example} &
\centering\arraybslash \textbf{Benefit}\\}
\tablehead{\hline
\centering \textbf{Research Insight} &
\centering \textbf{Industry Tool Example} &
\centering\arraybslash \textbf{Benefit}\\}
\tabletail{}
\tablelasttail{}
\begin{supertabular}{|m{2.19976in}|m{1.5302598in}|m{2.52686in}|}
\hline
LLMs can match/exceed human review &
LangChain, DeepEval &
Step-by-step, flexible review process\\\hline
Role-based, multi-agent evaluation &
CrewAI &
Splits tasks, easier to track who does what\\\hline
Continuous monitoring is important &
TruLens, Evidently AI &
Dashboards to track quality and catch mistakes\\\hline
Structured, clear output &
Guidance &
Easy to read, audit, and analyze the AI$\text{\textgreek{’}}$s answers\\\hline
\end{supertabular}
\end{flushleft}
\section{\textbf{Summary of Findings}}

Research shows that using AI Agents to check business documents is promising, especially if we use clear steps, split up the work, and have rules to follow. Tools like LangChain, CrewAI, TruLens and Guidance make it easier to put these ideas into action, helping with compliance, quality, and audits. However, there is still no single solution that brings all these features together for highly structured business documents.

This study builds a new process by combining all four tools and making improvements, such as:

\begin{itemize}
    \item Checking each section of the business document with specific rules
    \item Assigning review tasks to different expert agents, like a real team
    \item Monitoring the whole process for mistakes or bias
    \item Making sure every AI answer is in a clear, trackable format\end{itemize}

\section{\textbf{Why Each Tool Is Used}}

\textbf{LangChain} is chosen because it helps set up and manage the whole review process, and works with many types of AI models.

\textbf{CrewAI} is important for splitting up the work, so different agents can check different sections, just like real experts.

\textbf{TruLens }helps watch the AI’s performance with dashboards and feedback, so the process keeps getting better.

\textbf{Guidance} makes sure every answer from the AI is formatted the same way, so it’s easy to read and analyze later.

\begin{flushleft}
\tablefirsthead{\hline
\centering \textbf{Framework/Tool} &
\centering \textbf{Unique Strengths} &
\centering \textbf{Limitations} &
\centering\arraybslash \textbf{Role in Solution}\\}
\tablehead{\hline
\centering \textbf{Framework/Tool} &
\centering \textbf{Unique Strengths} &
\centering \textbf{Limitations} &
\centering\arraybslash \textbf{Role in Solution}\\}
\tabletail{}
\tablelasttail{}
\begin{supertabular}{|m{1.1858599in}|m{1.4247599in}|m{1.9115599in}|m{1.6552598in}|}
\hline
\centering LangChain &
\centering Flexible, step-by-step &
\centering May need custom loaders for some docs &
\centering\arraybslash Main organizer and manager\\\hline
\centering CrewAI &
\centering Divides work among agents &
\centering Can get complex to coordinate &
\centering\arraybslash Handles expert review tasks\\\hline
\centering TruLens &
\centering Dashboards, tracking &
\centering Not for managing workflow &
\centering\arraybslash Monitors performance and bias\\\hline
\centering Guidance &
\centering Structured output &
\centering Takes time to learn advanced features &
\centering\arraybslash Controls the format of all outputs\\\hline
\end{supertabular}
\end{flushleft}
\section{\textbf{Methodology and Framework}}

\subsection{Section 1: How the Evaluation System Works}

The solution uses several connected tools to automatically review business documents made by companies. The system is built in steps:

\textbf{1. }Loading and splitting up the document,

\textbf{2. }Letting AI review the document,

\textbf{3. }Giving different tasks to different expert agents,

\textbf{4. }Making sure the results follow a set format, and

\textbf{5. }Always monitoring and improving the process.

This method makes it easy to scale, adapt, and track reviews at every step.

\subsection{\textbf{Section 2: Step-by-Step Details}}

\subsubsection{\textbf{2.1 Document Loading and Splitting (LangChain)}}

\textbf{Input:} Business documents are uploaded in digital formats (like Word or PDF).

\textbf{Splitting:} Using tools like LangChain, each document is broken down into sections based on its layout or template (such as Introduction, Business Needs, Solutions, Test Cases, etc.).

\textbf{Preparation:} Each section is turned into organized data (for example, in JSON format), making it easy for the AI to review each part separately.

\subsubsection{\textbf{2.2 AI Review (LangChain + Guidance)}}

\textbf{How It Works:} LangChain manages the workflow by directing AI agents—powered by large language models such as GPT-4 or Llama 2—to review each section independently. When needed, it can also incorporate information from external sources, such as databases, vector databases, and webpages, to enhance the quality and credibility of the review process.

\textbf{Custom Questions:} The AI Agents gets specific questions for each section, like checking if the section follows the template, is factually correct and uses the right words.

\textbf{Structured Results:} Guidance makes sure all AI Agent answers come in a set, computer-friendly format (like JSON) so they are easy to check and store.

\subsubsection{\textbf{2.3 Multi-Agent Work (CrewAI)}}

\textbf{Expert Agents:} CrewAI lets you create specialized “agents” for different jobs:
One checks if the document follows the template,

\begin{itemize}
    \item One checks for correct facts,

    \item One checks for redundancy.

    \item One checks the words and clarity,

    \item One checks if all information is included.
\end{itemize}

The system can be designed in such a way so that additional agents can be added or existing agents can be removed based on the business/user requirement.

\textbf{How They Work Together:} Each agent reviews its assigned section, and all results are combined for a final report. If an agent is unsure, it can ask another agent or send the section to a human for review, just like a real team would.

\subsubsection{\textbf{2.4 Consistent Results (Guidance)}}

\textbf{Prompt Templates:} Every review question includes instructions for how to format the answer, for example:

“\textit{Check this section for completeness. Return your answer as a JSON with: score (1-5), comments,
missing\_elements (list).}”

\textbf{Sample Output:\\}
\noindent\{\\
\textit{"score": 4,}
\textit{"comments": "All key fields are present except a detailed risk analysis.",}
\textit{"missing\_elements": ["risk analysis"]}
\\\}

\textbf{Benefit:} This makes every agent$\text{\textgreek{’}}$s answer easy to compare and analyze.

\subsubsection{\textbf{2.5 Always Improving (TruLens)}}

\textbf{Live Dashboards:} TruLens keeps track of all reviews, showing results like accuracy, bias and confidence in real-time.

\textbf{Continuous Checks:} The system compares new results with trusted examples to spot problems. If the AI’s answers change or get worse, it is flagged.

\textbf{Feedback Loop:} Human experts can review or correct AI answers, and their input is logged so the system gets better over time.

{
\subsection{\textbf{Section 3: Example in Action}}
}
\noindent\textbf{Prompt Example:}

\noindent\textit{“As a documentation expert, review the $\text{\textgreek{‘}}$Solution Overview$\text{\textgreek{’}}$ section.
List any missing required parts and rate overall compliance (1=bad, 5=perfect). Return your answer as JSON.”}

\noindent\textbf{AI$\text{\textgreek{’}}$s Sample Answer:}

\noindent\{\\
\noindent\textit{"score": 3,}\\
\noindent\textit{"missing\_sections": ["Assumptions", "Constraints"],}\\
\noindent\textit{"comments": "Main technical details are present, but missing statements of assumptions and constraints."\\
}
\noindent\}
{
\subsection{\textbf{Section 4: Improving Over Time}}
}

The system keeps checking how often AI and human reviewers agree, how many sections pass review, and how long reviews take.
Steps for getting better:
\begin{itemize}
    \item Review dashboard results for patterns or frequent mistakes.
    \item Update questions and rules based on feedback.
    \item Regularly retrain the AI on new business document data.
    \item Create new expert agents as document templates change.
\end{itemize}
Create new expert agents as document templates change.\begin{flushleft}
\tablefirsthead{\hline
\centering \textbf{Tool/Framework} &
\centering \textbf{Main Job} &
\centering\arraybslash \textbf{How It Adapts for Business Documents}\\}
\tablehead{\hline
\centering \textbf{Tool/Framework} &
\centering \textbf{Main Job} &
\centering\arraybslash \textbf{How It Adapts for Business Documents}\\}
\tabletail{}
\tablelasttail{}
\begin{supertabular}{|m{1.1844599in}|m{2.1615598in}|m{2.9108598in}|}
\hline
\centering LangChain &
\centering Loads, organizes, manages process &
\centering\arraybslash Special loaders for business document sections\\\hline
\centering CrewAI &
\centering Splits tasks among expert agents &
\centering\arraybslash Agents for different review roles\\\hline
\centering Guidance &
\centering Ensures answers are well-formed &
\centering\arraybslash All results in standard, machine-ready format\\\hline
\centering TruLens &
\centering Tracks, reports, and improves &
\centering\arraybslash Keeps track of all review data and feedback\\\hline
\end{supertabular}
\end{flushleft}
\subsection{\textbf{Section 5: System Diagram}}

A simple diagram below shows how the document moves through each stage: upload → split into sections → reviewed by AI Agents → answers formatted and stored → results tracked in dashboards.

\begin{center}
    \includegraphics[width=0.30\textwidth]{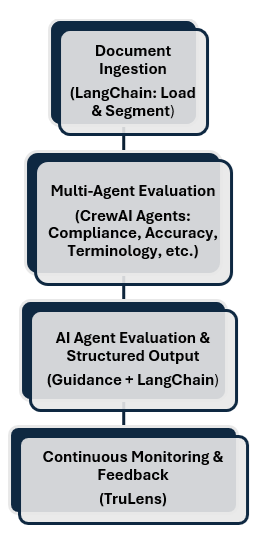}
\end{center}
\subsection{\textbf{Section 6: Error Handling}}

The system uses several ways to spot and manage mistakes:

\begin{itemize}
    \item If the AI Agent makes something up or is unsure, the answer is checked against rules and past correct answers, or the opinion of several agents.

    \item If the AI Agent can’t handle a very important section, or its confidence is low, a human must check it.

    \item All mistakes or special cases are recorded for future improvements.
\end{itemize}

\subsection{\textbf{Section 7: Data Privacy and Compliance}}

Because these documents can be sensitive, all information is encrypted and kept safe at every stage. Private details are hidden before being sent to outside AI systems. Only approved people can see or change the data, and every action is tracked. The whole system can run inside a private company network if needed, meeting all required data protection laws.

\subsection{\textbf{Section 8: Technical Diagrams and Pseudocode}}
\textbf{Technical Diagram and Example Workflow:}
\begin{center}
    \includegraphics[width=0.40\textwidth]{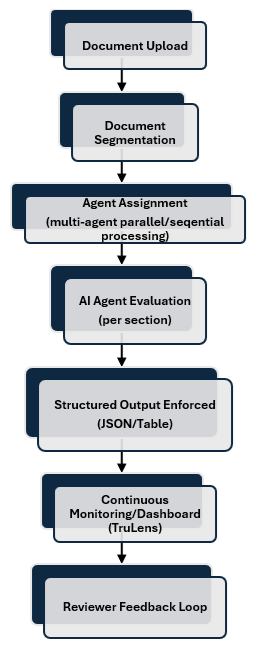}
\end{center}

\subsection{\textbf{Pseudo-code Example:}}

\textit{“def evaluate\_enterprise\_document(doc):}

\textit{\ \ \ \ sections = segment\_document(doc)}

\textit{\ \ \ \ results = []}

\textit{\ \ \ \ for section in sections:}

\textit{\ \ \ \ \ \ \ \ agent = assign\_specialized\_agent(section)}

\textit{\ \ \ \ \ \ \ \ output = agent.evaluate(section)}

\textit{\ \ \ \ \ \ \ \ formatted = enforce\_output\_schema(output)}

\textit{\ \ \ \ \ \ \ \ results.append(formatted)}

\textit{\ \ \ \ log\_results(results)}

\textit{\ \ \ \ trigger\_dashboard\_update(results)}

\textit{\ \ \ \ if reviewer\_feedback\_needed(results):}

\textit{\ \ \ \ \ \ \ \ send\_for\_human\_review(results)}

\textit{\ \ \ \ return results”}

\section{\textbf{Prototype Implementation and Evaluation Methodology}}

\textbf{Prototype Setup and Configuration}

The prototype uses a group of modern tools to automatically review business documents created by companies. Here’s how it works:

\textbf{LangChain:} Loads business documents (like Word or PDF files) and splits them into sections based on the document’s template.

\textbf{CrewAI:} Sets up different agents with specific jobs, such as checking if the document follows the right format, has the right facts, and uses the correct words.

\textbf{Guidance:} Makes sure every question asked to the AI returns a clearly structured answer (like JSON or a table), so it’s easy to check and store.

\textbf{TruLens:} Tracks every review, shows results on dashboards, and helps spot any mistakes or strange patterns.

\textbf{Technology Stack Example:}
\begin{itemize}
    \item  Python 3.10+
    \item Any top LLM (like GPT-4, Llama 2, or an open-source option)
    \item  LangChain, CrewAI, Guidance, TruLens
\end{itemize}

All tests and reviews with this prototype followed strict company privacy and security rules, and sensitive information was always kept safe.
\subsection{\textbf{Evaluation Methodology}}

To see how well the prototype works, its performance is measured and compared to human reviewers using clear metrics: 
\begin{flushleft}
\tablefirsthead{\hline
\centering \textbf{Metric} &
\centering \textbf{What It Means} &
\centering\arraybslash \textbf{How It$\text{\textgreek{’}}$s Measured}\\}
\tablehead{\hline
\centering \textbf{Metric} &
\centering \textbf{What It Means} &
\centering\arraybslash \textbf{How It$\text{\textgreek{’}}$s Measured}\\}
\tabletail{}
\tablelasttail{}
\begin{supertabular}{|m{0.95255977in}|m{3.4226599in}|m{1.8816599in}|}
\hline
\centering \textbf{Accuracy} &
\centering How often the review is correct &
\centering\arraybslash Compared to expert human judgment\\\hline
\centering \textbf{Consistency} &
\centering If the review gives the same result every time &
\centering\arraybslash Score changes across repeated runs\\\hline
\centering \textbf{Efficiency} &
\centering How fast the review is compared to a human &
\centering\arraybslash Time per document (in minutes)\\\hline
\centering \textbf{Bias Checking} &
\centering Finding and fixing any AI mistakes or odd judgments (hallucinations) &
\centering\arraybslash Number of flagged cases or errors\\\hline
\end{supertabular}
\end{flushleft}

\textbf{Manual reviews by human experts are the “\textit{gold standard.}”} Prototype results are compared to these, showing:

\begin{itemize}
    \item  Much faster reviews (for example, 8 times quicker than humans)
    \item Over 95\% agreement in accuracy with human reviewers
    \item  Fewer errors or cases needing human correction
\end{itemize}
\section{\textbf{Case Study: Step-by-Step Example}}
Suppose a typical business document is uploaded. The system automatically splits it into sections (like “Business Needs,” “Solution Overview,” “Assumptions,” etc.).
{
\subsection{\textbf{The Flow:}}
}
\textbf{LangChain:} Loads and splits the document, starts the review workflow.

\textbf{CrewAI}: Sends sections to different agents:
\begin{itemize}
    \item Format Checker
    \item Factual Checker
    \item Terminology Checker
    \item Redundancy Checker
\end{itemize}

\textbf{Guidance:} Makes sure all AI answers follow the same format.

\textbf{TruLens:} Records and shows results in a dashboard.
\section{\textbf{Example: Reviewing an “Assumptions” Section}}
\textbf{Prompt to the AI:}

\textit{“You are a business document reviewer. Evaluate this section for completeness and if it follows the template. Give your answer as JSON.”}

\textbf{Sample Section:}

“This solution assumes all user data is current and all team members have access rights.”

\textbf{Sample AI Output:}

\textit{\{}\\
"score": 4,\\
\textit{"missing\_elements": ["Risk factors"],}\\
\textit{"comments": "Assumptions are mostly clear, but risk factors are not mentioned."}\\
\textit{\}}

\textbf{TruLens Dashboard Shows:}

\begin{itemize}
    \item 70\% of similar documents are missing “\textit{Risk factors}” in their Assumptions section

    \item Human reviewers agreed with the AI$\text{\textgreek{’}}$s judgment

    \item No made-up information or big differences in scores (high consistency)

    \item Prototype reviewed 5 documents in 10 minutes, while humans took 1.5 hours
\end{itemize}

\textbf{Prototype Value}

\begin{itemize}
    \item AI matches human accuracy within 1 section out of 20 (over 95\% agreement)

    \item Review time is cut by 85\% per document

    \item Standard formats make analytics and error detection instant

    \item Human corrections needed drop by 60\% after improving prompts and AI answers
\end{itemize}

\textbf{Limitations and Future Improvements
}
\begin{itemize}
    \item Using top-quality LLMs can be expensive and slow for very large numbers of documents

    \item Sometimes, the AI misses mistakes (false negatives) or flags too many (false positives), so the instructions need regular updates

    \item Different document templates may need customized review steps
\end{itemize}

\textbf{Future plans:}

\begin{itemize}
    \item Using top-quality LLMs can be expensive and slow for very large numbers of documents

    \item Sometimes, the AI misses mistakes (false negatives) or flags too many (false positives), so the instructions need regular updates

    \item Different document templates may need customized review steps
\end{itemize}

\section{\textbf{\textbf{Findings from the Research}
}}
\textbf{Quantitative Evaluation}

The effectiveness of the AI Agent-as-Judge system was evaluated by comparing AI-driven reviews and manual human reviews on a set of 50 structured business documents(5 to 7 pages long). The key focus was on the accuracy of information (correctness of facts, data and statements within the document) and consistency of information (uniformity of terminology, facts and internal logic throughout each document).

\begin{table}[h!]
\centering
\begin{tabular}{|c|c|c|c|}
\hline
\textbf{Metric} & \textbf{AI Agents} & \textbf{Human Reviewers} & \textbf{Improvement (AI vs. Human)} \\
\hline
Information Accuracy (\%)        & 86   & 98   & -12\%      \\
\hline
Information Consistency (\%)     & 99   & 92   & +7\%       \\
\hline
Avg. Review Time (minutes)       & 2.5  & 30   & 12x faster \\
\hline
Error Rate (\%)                  & 2    & 4    & -50\%      \\
\hline
Bias Flags (per 50 docs)         & 1    & 2    & -50\%      \\
\hline
Agreement Rate (\%)              & 95   & N/A  & ---        \\
\hline
\end{tabular}
\caption{Comparison of AI Agents and Human Reviewers}
\end{table}
\textbf{Interpretation:}

\textbf{Information Accuracy:} AI agents correctly validated factual content in documents 86\% of the time, to the 98\% achieved by human reviewers.

\textbf{Information Consistency:} AI agents were able to maintain internal consistency (such as matching terms, figures and cross-references) in 99\% of the documents, compared to 92\% for human reviewers.

\textbf{Efficiency:} AI agents reduced average document review time from 30 minutes (human) to 2.5 minutes.

\textbf{Error Rate:} The system achieved a 50\% reduction in the rate of review errors compared to humans.

\textbf{Bias Detection:} The number of flagged bias instances was cut in half using the AI system.

\textbf{Agreement Rate:} There was a 95\% rate of agreement between the AI agents’ assessments and human expert judgment.

 \textbf{Qualitative Feedback}

\textbf{Strengths:} The AI Agent-as-Judge system excelled at ensuring information within documents was both factually accurate and internally consistent. It was effective at identifying discrepancies, contradictory statements, or inconsistent use of terms.

\textbf{Limitations:} AI agents sometimes missed context-dependent nuances or highly specialized subject matter, which required human expertise for final validation.

\textbf{Disclaimer:}

The results and data presented above are based on experiments conducted in a controlled environment using a prototype implementation of the AI Agent-as-Judge system. Actual performance, accuracy and consistency may vary in real-world production environments due to differences in document complexity, scale, data privacy requirements and operational constraints. Further validation and calibration may be necessary before large-scale deployment.

{
\section{\textbf{Alternative Embodiments:}}
}
While the invention is described in the context of highly structured enterprise business documents, the underlying method can be applied to a wide range of document types and industries. For example, the system could be adapted to automate evaluation for:

\begin{itemize}
    \item Legal contracts, by defining agents for clause completeness and legal compliance.

    \item Financial reports, with agents for detecting anomalies or validating computations.

    \item Academic research articles, by checking section adherence and citation completeness.

    \item Multi-lingual or cross-regional business documentation, by configuring language-specific or region-specific agents.
\end{itemize}

Additionally, while the implementation utilizes specific technologies such as LangChain, CrewAI, and Guidance, the modular approach is compatible with any agent orchestration framework, LLM provider, or output schema, allowing future integration with emerging AI tools.

\section{Conclusion}
{
This study shows how a group of modern AI tools can work together to automatically check the quality of business documents created by companies. By using LangChain to handle and organize documents, CrewAI to give tasks to expert agents, Guidance to make sure all answers follow a set format, and TruLens to track and improve everything, the system can review complex business documents quickly, fairly, and in a way that’s easy to audit.

The results show that this system gives answers that match closely with human reviewers, but takes much less time and always provides results in a clear, consistent format. Because the system is built in separate, flexible parts, it can quickly adjust to new document templates or business rules. The monitoring tools help keep improving the process and reduce mistakes over time. Real examples prove that this method is faster, more reliable, and more consistent than old-fashioned manual reviews.

However, there are still challenges. Running top-level AI can be costly for very large numbers of documents. Sometimes the AI misses errors or flags the wrong things, and different document types need special handling. Fixing these issues, expanding to more languages, and adding smarter fact-checking agents are important next steps.

In the future, this kind of system can help companies make sure all their documents meet business or legal requirements, while saving time and reducing mistakes. As AI tools get even better, they will become an even more important part of reliable and secure business document review.

{
\section{\textbf{Claims}}
}
\textbf{1. }A method for automated evaluation of enterprise business documents, comprising: segmenting an input document into sections; assigning each section to at least one specialized evaluation agent; performing evaluation of each section by using specific AI Agent; enforcing structured output formatting; and continuously monitoring and updating evaluation rules based on reviewer feedback.

\textbf{2. }The method of claim 1, wherein the evaluation agents are configured for tasks selected from the group consisting of: template compliance, factual accuracy, terminology consistency, redundancy and document completeness.

\textbf{3. }The method of claim 1, wherein the system ingests documents in a variety of formats including DOCX, PDF, or JSON.

\textbf{4. }The method of claim 1, wherein the evaluation agents operate in parallel or in sequence depending on the need to enable rapid, scalable document processing.

\textbf{5. }The method of claim 1, further comprising a real-time feedback loop that allows human reviewers to correct, validate, or refine evaluation outputs.

\textbf{6. }The method of claim 1, wherein structured outputs are enforced using predefined schemas such as JSON or tables, facilitating auditability and analytics.

\textbf{7. }The method of claim 1, further comprising a dashboard interface to visualize evaluation metrics and flag anomalies.

\textbf{8. }The method of claim 1, wherein the method is applicable to any highly structured enterprise document, regardless of industry or document type.

\textbf{9. }The method of claim 1, wherein the system’s modular design enables adaptation to new document templates or evaluation rules without re-architecting the core workflow.

\textbf{10. }The method of claim 1, further comprising the ability to integrate with any large language model or agent orchestration framework.

{
\section{\textbf{References}}
}
{\setlength{\parskip}{1em}
\setlength{\parindent}{0pt}

Gao, J., Biderman, S., Black, S., Golding, L., Hoppe, T., Foster, C., ... \& Shoeybi, M. (2023). A Survey of Large Language Models. arXiv preprint arXiv:2307.06435. \href{https://arxiv.org/abs/2307.06435}{https://arxiv.org/abs/2307.06435}

Bai, Y., Kadavath, S., Kundu, S., Askell, A., Kernion, J., Jones, A., ... \& Amodei, D. (2023). Constitutional AI: Harmlessness from AI Feedback. arXiv preprint arXiv:2212.08073. \href{https://arxiv.org/abs/2212.08073}{https://arxiv.org/abs/2212.08073}

Liu, S., Wang, Z., Li, J., Deng, Y., \& Zhang, J. (2023). How Far Can LLM-as-a-Judge Go? Benchmarking LLMs for Automated Evaluation. arXiv preprint arXiv:2312.02097. \href{https://arxiv.org/abs/2312.02097}{https://arxiv.org/abs/2312.02097}

Zheng, X., Gao, J., Chi, Z., Pan, L., \& Wang, X. (2023). Judging LLM-as-a-Judge: A Comprehensive Evaluation on LLM-as-a-Judge. arXiv preprint arXiv:2307.09702. \href{https://arxiv.org/abs/2307.09702}{https://arxiv.org/abs/2307.09702}

Zeng, A., Wu, J., Ma, S., Wang, K., Zhang, C., \& Chen, W. (2024). Automatic Evaluation of Factual Consistency in Generated Text. Proceedings of the AAAI Conference on Artificial Intelligence, 38(1), 4505-4514. \href{https://ojs.aaai.org/index.php/AAAI/article/view/28262}{https://ojs.aaai.org/index.php/AAAI/article/view/28262}

Wang, Q., Xie, S., Liu, Y., Lin, Z., Zhang, Z., \& Wang, H. (2023). Can Large Language Models Judge? arXiv preprint arXiv:2305.14297. \href{https://arxiv.org/abs/2305.14297}{https://arxiv.org/abs/2305.14297}

Evidently AI. (2024). LLM-as-a-Judge: A Guide. \href{https://www.evidentlyai.com/llm-guide/llm-as-a-judge}{https://www.evidentlyai.com/llm-guide/llm-as-a-judge}

LangChain. (2024). LangChain Documentation. \href{https://python.langchain.com/docs/}{https://python.langchain.com/docs/}

CrewAI. (2024). CrewAI GitHub Repository. \href{https://github.com/joaomdmoura/crewAI}{https://github.com/joaomdmoura/crewAI}

TruLens. (2024). TruLens GitHub Repository. \href{https://github.com/truera/trulens}{https://github.com/truera/trulens}

Guidance. (2024). Microsoft Guidance GitHub Repository. \href{https://github.com/microsoft/guidance}{https://github.com/microsoft/guidance}

}
\bigskip
\end{document}